\documentclass[11pt]{article}
\usepackage{geometry}
\usepackage{coling2020}
\usepackage{times}
\usepackage{url}
\usepackage{latexsym}
\usepackage{microtype}
\usepackage{graphicx}

\usepackage{float}
\usepackage{mdframed}

\colingfinalcopy 

\title{DSC IIT-ISM at SemEval-2020 Task 6: Boosting BERT with Dependencies for Definition Extraction}

\author{
\\
  \\
  \And
  Aadarsh Singh*, Priyanshu Kumar* and Aman Sinha \\
  \\
  Indian Institute of Technology (Indian School of Mines) Dhanbad, India \\
  {\tt \{aadarshsingh191198, kpriyanshu256, amansinha091\}@gmail.com} \\
  \And
  \\
  \\
  \\}
 
\date{}

\begin{document}
\maketitle

\begin{abstract}
  We explore the performance of Bidirectional Encoder Representations from Transformers (BERT) at definition extraction. We further propose a joint model of BERT and Text Level Graph Convolutional Network so as to incorporate dependencies into the model. Our proposed model produces better results than BERT and achieves comparable results to BERT with fine tuned language model in DeftEval (Task 6 of SemEval 2020), a shared task of classifying whether a sentence contains a definition or not (Subtask 1). 
\end{abstract}

\section{Introduction}
\label{intro}

%
%
\blfootnote{
    %
    %
    \hspace{-0.65cm}  
    \textbf{\textit{*}} Equal contribution.\\
    This work is licensed under a Creative Commons Attribution 4.0 International Licence. Licence details: \url{http://creativecommons.org/licenses/by/4.0/}
    %
    %
    %
    %
}

Definition Extraction from free text (DEFT) \cite{shirani2020semeval}  involves finding term definition pairs from free and semi-structured texts, especially those whose term-definition span crosses a sentence boundary and those which do not have a definition phrase. An example of a cross boundary sentence from the DEFT corpus is given below:
\vspace{0.3cm}
\begin{mdframed}
In doing so , $<$DEF$>$ monomers release water molecules as byproducts
$<$DEF$>$ (1). This type of reaction is known as $<$TERM$>$ dehydration
synthesis $<$TERM$>$ , which means “ $<$QUALIFIER$>$ to put together while losing water $<$QUALIFIER$>$ . ”(0)
\end{mdframed}
\vspace{0.3cm}
It can be observed that the definition (enclosed by DEF tags) and the corresponding terms (enclosed by TERM tags) are present in different sentences, thus increasing the difficulty of definition extraction. The task is thus relatively different and complex from the conventional definition extraction (DE) task in which a definition could be broken into the following sub-parts:

\begin{enumerate}
	\item The DEFINIENDUM field (DF) i.e., the word being defined.
	\item The DEFINITOR field (VF) i.e., the verb phrase used to introduce the definition.
	\item The DEFINIENS field (GF) i.e., genus phrase or the hypernym.
	\item The REST field (RF) i.e., additional clauses that help to distinguish the definiendum from its genus.
\end{enumerate}

and thus easily be captured by common verb phrases (DEFINITOR) like “means”, “refers to”, “is”, etc. These kind of conventional definitions could easily be tagged as follows:

\vspace{0.3cm}

\begin{mdframed}

$<$DF$>$Photosynthesis$<$/DF$>$ $<$VF$>$is$<$/VF$>$ $<$GF$>$the process $<$/GF$>$ $<$RF$>$by which green plants manufacture food. $<$/RF$>$
\end{mdframed}
\vspace{0.3cm}

In the example presented above, the definition (REST field) and the corresponding term (DEFINIENDUM field) are present in the same sentence. Moreover, the presence of DEFINITOR(s) in the text also eases the task of extracting such definitions. 

\newpage



\noindent
The order of the tags in the WCL Corpus \cite{navigli2010annotated}, one of the conventional corpus for Definition Extraction task, is predefined viz. DEFINIENDUM, DEFINITOR, DEFINIENS, REST. On the other hand, there is no such predefined order for the DEFT corpus regarding the occurrence of BIO Tags. This absence of order makes it difficult for the models to identify the definitional sentences in the DEFT corpus and then associate the words present in the definitions with the appropriate tags.

Another observation is the variation in the pattern of the definitional sentences in the DEFT corpus due to the presence of the heterogeneous distribution of tags. Some tags like ‘O’,‘I-Term’,‘I-Definition’, etc. are very frequent whereas tags such as ‘Alias-Term’, ‘Secondary-Definition’ are rarely seen. This imbalance in dataset causes difficulty in finding a structure for the definition and hence makes the DEFT corpus relatively more complex when compared to conventional corpora like WCL. 

\begin{figure}[H]
\center
\includegraphics[width=\textwidth]{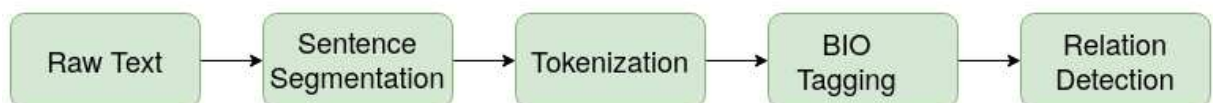}
\caption{Pipeline for Definition Extraction}
\label{fig:pipeline}
\end{figure}

Thus, given the complexity of the definitions present in the DEFT corpus, the whole pipeline for the extraction of meaningful term-definition pairs from free text can be restructured as presented in Figure \ref{fig:pipeline}.

\noindent
Since the corpus is already in the form of sentences, sentence segmentation is not required. For tagging tokens, authors of DEFT Corpus \cite{spala2019deft} have defined a new annotation scheme, a part of which has been touched upon in the introduction section. Moreover, it can be found in the corresponding paper. 



In this paper, we present the approach used for the task ``DeftEval: Extracting term-definition pairs in free text" of SemEval 2020. The subtasks are as follows: 

\begin{enumerate}
\item Subtask 1 - Sentence Classification
\item Subtask 2 - Sequence Labeling
\item Subtask 3 - Relation Classification

\end{enumerate}

Our presented methods for the sentence classification subtask (Subtask 1) revolve around the Transformer \cite{vaswani2017attention} based model, Bidirectional Encoder Representations from Transformers (BERT) \cite{devlin2018bert}. Experiments \footnote{Corresponding submissions have been made under the username of {\tt mler} on Codalab} have been done to improve the results and in the process we have come up with improvised architectures.

The rest of the paper is organized as follows: Related work with its limitations has been discussed in Section \ref{sec:relwork}, followed by a description of the data used in Section \ref{sec:data}. The proposed methods have been elaborated in Section \ref{sec:methods} \footnote{Source code available at \url{https://github.com/dsciitism/SemEval-2020-Task-6} }. Section \ref{sec:results} and \ref{sec:error} contains the results and error analysis respectively. Section \ref{sec:end} concludes the paper and also includes the possible future work.

\section{Related Work}
\label{sec:relwork}
The very first attempt at DE was by \newcite{kobylinski2008definition}. They used a “Balanced Random Forest classifier” to classify definitions so as to handle the imbalance of the dataset. A major contribution to DE was by \newcite{navigli2010learning}, by using a generalization of word lattices to model textual definitions in the form of Definiendum, Definitor, Definiens and Rest. However, this approach is unable to generalise definitions, especially the ones that defy the conventional semantics i.e. common patterns found in definitional sentences.

\noindent
Many existing works take advantage of linguistic features like syntactic dependencies. \newcite{jin2013mining} built DefMiner which is a supervised sequence labelling system using shallow parsing and dependency features. Training CRF (Conditional Random Fields) with lexical, terminological, and structural features extracted from data has also been tried by \newcite{anke2013towards}. The use of linguistic features lead to promising results and observations. \newcite{espinosa2014applying} took advantage of the subtrees in the dependency parsing of a sentence. Feature representation of sentences were created using these subtrees. \newcite{espinosa2015definition} incorporated both linguistic and semantic features of sentence for training classifiers. SensEmbed is used to reveal the semantic compactness of definition containing sentences.

With the advent of the deep learning era, the first attempt at DE using deep learning was by \newcite{li2016definition}. Feature representation of sentences were learned using Long Short Term Memory (LSTM) cells. Text preprocessing such as replacing some selected words with POS tags, were carried out thus obtaining brilliant results. \newcite{anke2018syntactically} developed Syntactically Aware Neural Networks by incorporating syntactic information (syntactic dependencies and dependency labels) along with the text sentences as input. With the help of pre-trained word embeddings, convolutional filters and Bi-LSTM cells, the model was capable of extracting both short range and long range dependencies from the text data.

In the most recent attempt at Definition Extraction, \newcite{veyseh2019joint} proposed a multi-task model to perform sentence classification and sequence labelling simultaneously. The model took advantage of the entire syntactic dependency tree rather than just dependencies, thus yielding state of the art results on the WCL dataset.

\section{Dataset}
\label{sec:data}

For the purpose of training, evaluation and testing, the DEFT corpus has been used. Not only is the corpus significantly larger than the previously available corpora in the field of Definition Extraction, but the dataset also contains definitions from complex, human-annotated data across a variety of topics and from both free (textbook) and semi-structured (legal document) language. Table \ref{dataset} shows the statistics of the major datasets available in the field of DE namely, WCL, W00 \cite{jin2013mining} and DEFT.

\begin{table}[htpb]
\centering
\begin{tabular}{|l|r|c|}
\hline \bf Dataset & \bf No. of positive annotations & \bf Size(in sentences) \\ \hline
WCL & 1,871 & 4,718 \\ \hline
W00 & 731 & 2,185 \\ \hline
DEFT & 11,004 & 23,746 \\ \hline
\end{tabular}
\caption{\label{dataset} Dataset statistics }
\end{table}

By complicated structure, we mean that the corpus does not contain simple sentences of the form “X is a Y” as is the case in most of the definitions present in the WCL dataset. Instead, roughly 50\% of term-definition pairs in the dataset appear across sentence boundaries or with an otherwise complex structure (e.g., containing secondary information, containing ambiguous references to previously stated terms or definitions) whereby the relationship between a term and definition requires more deduction than finding a definition verb phrase.

\section{Methods}
\label{sec:methods}

\subsection{Data Preprocessing}

Unlike conventional approaches for text classification (Tf-Idf, Bag of Words Model, etc.), deep learning approaches require minimal preprocessing. It is often believed that preprocessing leads to the loss of information. So, for preprocessing the we have only performed the following two steps:
\begin{enumerate}
\item Removal of leading line numbers: Some of the sentences in the data have a leading line number. It has been removed since it is irrelevant to the content of the sentence and acts as noise.

\item Addition of the subject token: It has been found in some cases of our experiments that adding the subject token i.e. the subject of the textbook from which the sentence to be classified is picked, helps improve the results.
\end{enumerate}

\subsection{Model Architectures}

Initially, we started off by using the approach of Syntactically Aware Neural Networks. However, owing to the complexity of the DEFT corpus, the model was unable to perform well. Hence, we shifted towards using larger pretrained models such as Transformers. We concentrate on the performance of BERT since it has achieved the State of The Art results in many NLP tasks with  limited  fine-tuning  on task-specific training data. Moreover, \newcite{10.1007/978-981-15-6318-8_13} have shown that BERT implicitly captures syntactic dependencies, which play an essential role in Definition Extraction.

We now describe the three approaches that we have implemented and submitted to the shared task. 

\begin{enumerate}

\item {\bf BERT} : BERT, based on the Transformer architecture, consists of multi-attention heads which apply sequence-to-sequence transformation on the input text sequence. 
BERT incorporates the following practices for training 
(a) learn to predict a masked token using the left and right context of the text sequence (Masked Language Model) 
(b) learn to predict whether two sentences occur in continuation or not (Next Sentence Prediction)

For our experiments, we use the BERT (base-cased) made publicly available by Huggingface \cite{wolf2019transformers}. It consists of 12 hidden layers in the encoder of the Transformer. The encoder outputs a feature vector of 768 dimensions. We select the cased model (pre-trained on cased English text) because lower casing the data leads to loss of information. The representation corresponding to the CLS token is fed through two feed-forward linear layers, thus giving two values corresponding to the logits of the two classes. We train the model for 5 epochs using AdamW \cite{DBLP:journals/corr/abs-1711-05101} optimizer with a learning rate of 2e-5.

\item {\bf BERT with fine tuned language model} : This model is the same as the first one, except we fine tune the Masked Language Model of BERT using the training data. This helps BERT to understand the context of the corpus in a better manner. Results show that fine tuning the language model of neural networks gives improved results \cite{howard2018universal}. 

To fine tune the LM, we take help of Huggingface's \textit{run\_language\_modeling.py} utility\footnote{\url{https://github.com/huggingface/transformers/tree/master/examples/language-modeling}}. The other hyperparameter settings remain the same as the first model.

\begin{figure}[ht]
\center
\includegraphics[width=0.55\textwidth]{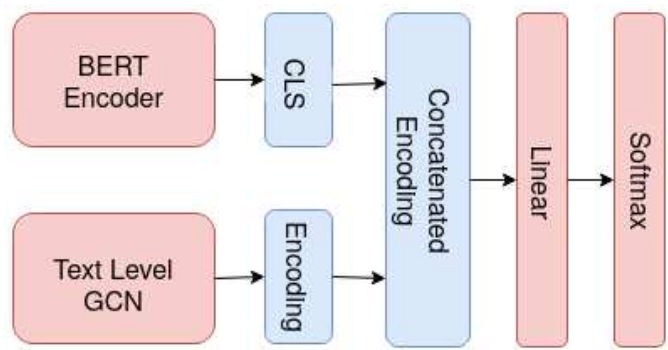}
\caption{Joint model of BERT and Text Level GCN}
\label{fig:hyb}
\end{figure}

\item {\bf Joint model of BERT and Text Level GCN} : The above models do not take into account linguistic features along with the text data. Researchers have obtained superior results by incorporating linguistic information into the models. However, creating such features requires extra effort and involves the use of dependency parsers, which may lead to error propagation (error in extracting linguistic information will lead to feeding of erroneous information to the model). Graph Convolutional Networks (GCN) \cite{kipf2016semi} have yielded superior performance on graph-structured data. Since text can also be represented as graph data (for example syntactic dependency trees), we use the feature representation obtained from a Text Level Graph Convolutional Network \cite{huang2019text} along with the feature representation obtained from BERT, to solve our task. As shown in Figure \ref{fig:hyb}, both the representations are concatenated and a linear layer is used to output the logits of the two classes.

\begin{figure}[ht]
\center
\includegraphics[width=0.35\textwidth]{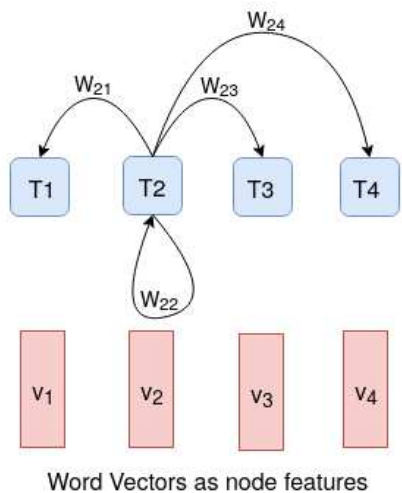}
\caption{Inner Working of Text Level Graph Convolutional Networks}
\label{fig:gcn}
\end{figure}

In a GCN, data is represented as graph(s) and each node of the graph is represented by a set of attributes. A Text Level GCN represents a text sequence as a graph. Each token of the sentence represents a node and has its corresponding word vector as attributes. A token has an edge between its n-grams to the left and right of it (n is also referred as window size). The greater the value of n, greater is the range of dependencies that are captured. Figure \ref{fig:gcn} shows the corresponding edges for a token (T2) of a text sequence with 4 tokens, considering a window size of 2.

The weights of the edges are learned through training. The weighted representations (product of edge weight and word vector) is sent to its neighbours by each node. Each node gathers information from its neighbouring nodes, and uses it to update its own representation (i.e. attributes) and edge weights. The values of the edge weights represent the significance of the dependency between a token and one of its n-grams. Thus, the model understands the difference between strong and weak dependencies without any human intervention. 

The language model of the BERT component of the model is not fine tuned. We use the publicly available pre-trained weights. The Text Level GCN component uses pre-trained GloVe word embeddings \cite{pennington2014glove} as node attributes. We use a window size of 5 in our experiments. The model is trained for 5 epochs using AdamW with learning rate 2e-5.

\end{enumerate}

\section{Results}
\label{sec:results}

We evaluate the performance of the three models on the validation set and the test set. The evaluation metric is F1 score on the positive class. The submissions of the test set were done using a 10-fold cross validation so as to increase the robustness of the results. The results are presented in Table \ref{results}.

\begin{table}[htbp]
\centering
\begin{tabular}{|l|l|l|}
\hline \bf Model & \bf Validation Set & \bf Test Set \\ \hline
BERT & 0.74 & 0.731 \\ \hline
BERT with fine tuned language model & 0.768 & \textbf{0.775} \\ \hline
Joint model of BERT \& GCN & \textbf{0.781} & 0.758 \\ \hline
\end{tabular}

\caption{\label{results} Comparison of results of the mentioned models }
\end{table}

The results show that fine tuning the language model of BERT leads to improved results. The fine tuned BERT achieves the greatest F1 score on the test set among the 3 models. The reason for such a performance boost can be attributed to the structure of the DEFT corpus. The corpus consists of long continuous segments from textbooks and legal documents. Since there is a lot of text in the same context, fine-tuning the language model helps BERT to get a better understanding of the domain of the corpus.

It is also evident from Table \ref{results} that the inclusion of Text Level GCN along with BERT is also beneficial. The Text Level GCN component of the joint model is capturing some extra information from the text data which is boosting the performance.

\section{Error Analysis}
\label{sec:error}

We analyze the predictions on the validation set. For the predictions of all the models, we calculate the binary cross entropy loss for each example and then sort the examples as per descending loss values. We examine the top commonly mis-classified examples by each of the 3 models (in Table \ref{table:error}).

\begin{table}[htbp]
\centering
\begin{tabular}{|p{12cm}| l|} 
\hline \bf Sentence & \bf True Label \\ \hline
”United States v. Miller , 307 U.S. 174 ( 1939 ) . & 1 \\ \hline
Pathogens include bacteria , protists , fungi and other infectious organisms . & 1 \\ \hline
Toll goods are available to many people , and many people can make use of them , but only if they can pay the price . & 1 \\ \hline
\end{tabular}
\caption{\label{table:error} Common highly mis-classified sentences }
\end{table}

We observe that majority of these examples belong to the cross sentence definition scenario as mentioned in Section \ref{intro}. This implies that models are struggling to extract cross-sentence definitions. The models lack a context to the sentence to be classified, because of which it is unable to classify definitions that are covered in more than one sentence. To the human reader, the predictions of most of mis-classified sentences would appear correct. However, they are incorrect according to DEFT corpus.

An additional reason for the errors is the incorrect/ambiguous labelling of the dataset. For example, consider the sentence ‘”United States v. Miller , 307 U.S. 174 ( 1939 ) .' in table \ref{table:error}. The sentence is not a definition but due to the inverted comma (”) of the trailing sentence (which was a definition), this sentence was also labelled a definition. Similarly, the label for the sentence “Pathogens include bacteria , protists , fungi and other infectious organisms ." is debatable as it seems more of a description than a definition. Last but not the least, “Toll goods are available to many people , and many people can make use of them , but only if they can pay the price ." is clearly not a definition.


\section{Conclusion and Future Work}
\label{sec:end}

We study the performance of BERT on the DEFT corpus and tried to boost BERT with explicit linguistic information (in the form of dependencies) using a Text Level GCN model. For further improvements, we can use BERT with a fine-tuned language model as the component of the joint model. The models can also make use of a context in the form of the sentence preceding the sentence to be classified. The incorporation of a context can help models overcome the problem of misclassifying cross-sentence definitions. 

\bibliographystyle{coling}
\bibliography{semeval2020}

\end{document}